\documentclass[runningheads]{llncs}

\usepackage{eccv}

\usepackage{eccvabbrv}

\usepackage{graphicx}
\usepackage{booktabs}
\usepackage{amsmath,amssymb} % define this before the line numbering.
\usepackage{color}
\usepackage{adjustbox}
\usepackage{multirow}

\usepackage[accsupp]{axessibility}  % Improves PDF readability for those with disabilities.

\usepackage{hyperref}

\usepackage{orcidlink}

\begin{document}

\title{CIPER: A Unified Framework for Cross-view Image-retrieval and Pose-estimation
} 

\titlerunning{CIPER}

% \author{Yurim Jeon\inst{1}\and
% Dongseong Seo\inst{1} \and
% Seung-Woo Seo\inst{1}}
\author{Yurim Jeon\inst{1}$^{\dagger}$\and
Dongseong Seo\inst{1}$^{\dagger}$ \and
Seung-Woo Seo\inst{1}}

\authorrunning{Y.Jeon et al.}

\institute{Seoul National University, Seoul, Republic of Korea \and
\email{\{fabioisyo01,sorld0603,sseo\}@snu.ac.kr}}

\footnotetext{$^{\dagger}$ These authors contributed equally to this work.}

\maketitle

\begin{abstract}
% Cross-view image geo-localization involves estimating the location of a query ground image within a reference aerial image database. This task has gained popularity with the widespread use of mobile cameras and advancements in aerial imaging technology. Previous studies employed two approaches to address this problem: image retrieval, which broadens the search range but sacrifices accuracy, and pose estimation, which offers high positional accuracy but within a limited search range. Our proposed solution, Cross-view Image-retrieval and Pose-estimation transformER (CIPER), unifies these two approaches to provide a versatile cross-view image geo-localization solution that includes a wide search range and high positional accuracy. We evaluated the proposed method on large-scale datasets for both image retrieval and pose estimation. The results demonstrate the efficacy of the proposed method in practical settings. Code will be released.

Cross-view geo-localization aims to estimate the geographic location of a ground image using a reference database of aerial images. Existing approaches typically address this problem through either large-scale image retrieval or precise pose estimation. While retrieval-based methods enable wide-area search, their localization accuracy is limited by database resolution. In contrast, pose estimation methods achieve high accuracy but are constrained to a narrow search space. However, simply cascading these disjoint pipelines often suffers from error propagation and inconsistent feature representations. To overcome these limitations, we formulate cross-view geo-localization as a unified problem that simultaneously requires city-scale retrieval and precise 3-DoF pose estimation. To address this challenge, we propose CIPER (Cross-view Image-retrieval and Pose-estimation transformER), a unified framework that jointly performs both tasks within a single architecture, promoting mutually beneficial feature learning. CIPER employs a shared transformer encoder with task-specific tokens to disentangle global retrieval features and spatial localization cues. To mitigate the significant domain gap between ground and aerial images, we propose a two-way transformer pose decoder that leverages ground features as spatial queries to perform bidirectional cross-attention for robust cross-view alignment. Furthermore, a set prediction strategy is adopted for stable 3-DoF regression under a unified multi-task learning objective. Extensive experiments on large-scale datasets, including VIGOR, KITTI, and Ford Multi-AV, demonstrate that the proposed method achieves reliable and competitive performance, particularly under limited field-of-view and arbitrary orientation conditions. These results validate CIPER as a versatile and robust baseline for practical cross-view localization, providing a reliable foundation for future research in unified architectures. Code is available at \url{https://github.com/yurimjeon1892/CIPER}.

% TODO 260127
\end{abstract}

\section{Introduction}

We live in an era in which countless photos are uploaded to the Internet from mobile devices every second. As a result, more ground images (i.e., images captured from the ground) are being generated more than ever before. Simultaneously, accessibility to aerial images has improved through services like Google Maps. This trend has led to an increase in tasks that involve both ground and aerial images. Among these, cross-view geo-localization, which estimates the location where a query ground image was captured from a reference database of aerial images, is prominent. Because of its high potential for practical applications, this task is particularly noteworthy. It can be employed in GPS-denied environments to establish the datum of dead reckoning or can assist vulnerable GPS receivers in applications such as robot navigation and autonomous driving~\cite{yan2022crossloc}.

Two primary approaches are used to solve the cross-view geo-localization problem. The first approach is image retrieval~\cite{hu2018cvm}, \cite{shi2019spatial}, \cite{zhu2021vigor}, \cite{zhu2022transgeo}. Image retrieval approach queries a ground image to a database of GPS-tagged aerial images and estimates the location based on the GPS tags of matching aerial images. It has the advantage of a broad search range that is as wide as the coverage of the aerial images in the database. However, the accuracy of localization is dependent on the sampling frequency of the aerial images. Another approach that addresses this problem is pose estimation~\cite{xia2022visual}, \cite{shi2022beyond}, \cite{lentsch2023slicematch}, \cite{shi2023boosting} \cite{wang2023pureACL}. This approach ensures high localization accuracy by estimating the location as a three-degree-of-freedom (3-DoF) pose. However, prior to pose estimation, the matching aerial image for the queried ground image must be found. Therefore, under pose estimation alone, this approach is constrained to the range of a single aerial image.

\begin{figure*}[t]
    \begin{center}
    \includegraphics[width=0.6\linewidth]{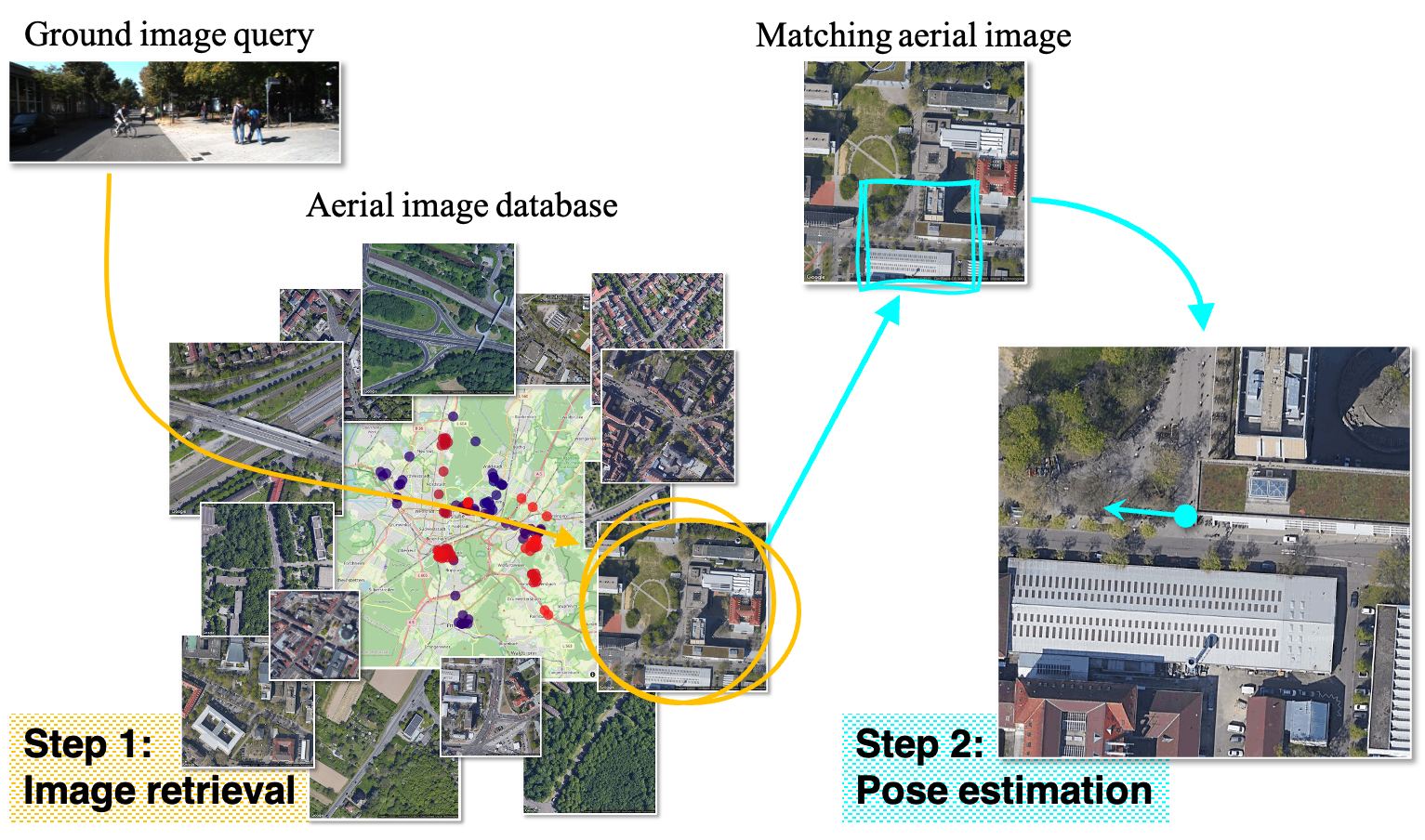}
    \end{center}
    \vspace*{-0.5cm}
    \caption{\textbf{Our scenario} 
    A query ground image and city-scale aerial image database are given. The goal is to estimate the location with a 3-DoF pose where the query was captured.
    }
    \label{fig:int_scenario}
\end{figure*}

% These two independent approaches have simplified the task of cross-view image geo-localization, achieving notable success in the fields of image retrieval and pose estimation. However, it also has limitations in real-world applications, as previously described. Here, we present a practical cross-view image geo-localization scenario that challenges the aforementioned approaches. Our scenario is described as follows. Given a query ground image and a city-scale aerial image database, the goal is to estimate the location with a 3-DoF pose where the query was captured. This scenario simultaneously requires both a wide search range and high positional accuracy. To address this scenario, the solution first searches for candidates in a city-scale database relevant to the given query image (image retrieval). It then estimates the 3-DoF pose corresponding to the location of the query within the candidate pool of aerial images (pose estimation). Essentially, a unified approach that integrates the previous approaches is required. Our scenario is shown in Fig.~\ref{fig:int_scenario}.
Two independent approaches—image retrieval and pose estimation—have significantly simplified the task of cross-view geo-localization and achieved notable success in their respective domains. However, applying these approaches separately exhibits critical limitations in real-world applications. To address this gap, we define a practical cross-view geo-localization scenario: given a query ground image and a city-scale aerial image database, the objective is to estimate the precise 3-DoF pose of the location where the query image was captured. This scenario requires both a wide search space across the entire city and high positional accuracy. Simply cascading disjoint pipelines for this purpose—performing image retrieval followed by an independent pose estimation step—leads to redundant feature extraction and computational inefficiency, as each module processes the large aerial database separately. Therefore, a unified framework that evaluates both global context for retrieval and local spatial cues for pose estimation within a shared feature space is essential. The overall scenario is illustrated in Fig.~\ref{fig:int_scenario}.

%

\begin{figure*}[t]
    \begin{center}
    \includegraphics[width=1\linewidth]{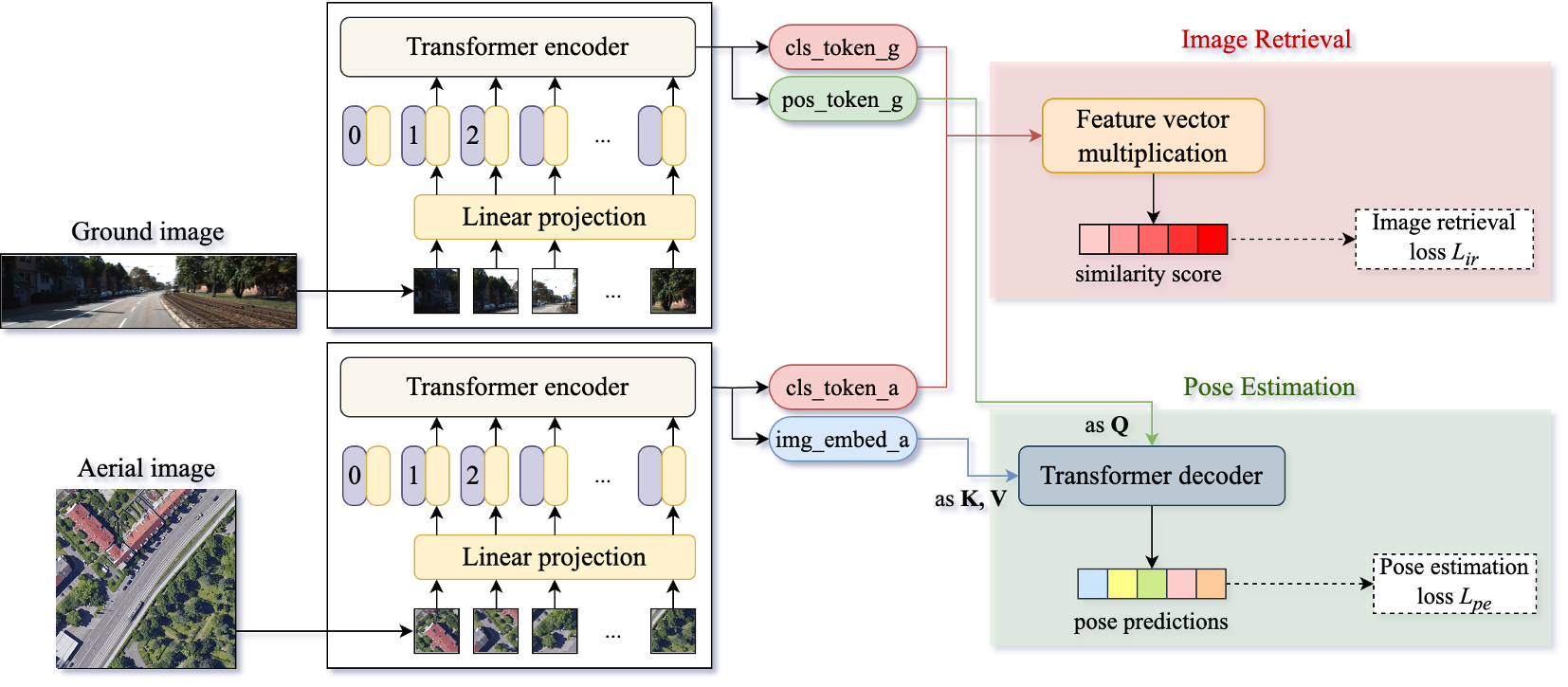}
    \end{center}
    \vspace*{-0.35cm}
    \caption{\textbf{Overview of the proposed method.} 
    Ground and aerial images are processed by a shared image encoder, generating multi-purpose features for both image retrieval and pose estimation. The \textit{cls\_token} is utilized for city-scale image retrieval via similarity score computation. The top-matched aerial images then serve as localized context for precise pose estimation without feature re-extraction. During pose estimation, the \textit{pos\_token\_g} acts as the spatial query for the pose decoder, performing bidirectional cross-attention with the aerial features (\textit{img\_embed\_a}). Consequently, the 3-DoF pose of the ground image relative to the aerial image is directly estimated.}
    \label{fig:mtd_overivew}
\end{figure*}

% In this study, we propose a Cross-view Image-retrieval and Pose-estimation transformER (CIPER) as a solution to the aforementioned scenario. CIPER consists of two parts: image encoder, and pose decoder. The image encoder generates multi-purpose features for both image retrieval and pose estimation, optimizing memory usage. The pose decoder, which is based on a two-way transformer, enables one-shot high-precision pose estimation without the need for an iterative process. The structure of CIPER is illustrated in Fig.~\ref{fig:mtd_overivew}. 
% TODO 260127
To address the unified cross-view geo-localization problem, we propose Cross-view Image-retrieval and Pose-estimation transformER (CIPER), a single network that jointly performs large-scale image retrieval and precise 3-DoF pose estimation within a unified framework. Unlike conventional pipelines that treat retrieval and pose estimation as independent stages, CIPER integrates both tasks through a shared image encoder and a dedicated pose decoder. The encoder extracts both global and local features using a transformer backbone augmented with learnable class and pose tokens; the class token is optimized for discriminative retrieval, while the pose token preserves essential spatial cues for localization. To bridge the severe domain gap between ground and aerial images, we introduce a two-way transformer–based pose decoder. This decoder treats ground features as spatial queries to perform bidirectional cross-attention with aerial features, establishing a robust cross-view alignment mechanism that is highly resilient to viewpoint variations and limited fields of view. Furthermore, by adopting a set prediction strategy for stable 3-DoF regression, our pose decoder enables direct and stable high-precision estimation, eliminating the dependency on the complex iterative optimization processes typical of existing methods. Through this unified design, CIPER enables scalable city-level search while maintaining high positional accuracy, overcoming the limitations of disjoint cross-view localization pipelines. The structure of CIPER is illustrated in Fig.~\ref{fig:mtd_overivew}.

We evaluate the proposed method on both image retrieval and pose estimation tasks using large-scale datasets (VIGOR~\cite{zhu2021vigor}, KITTI~\cite{geiger2013vision}, and Ford multi-AV~\cite{agarwal2020ford}). The results demonstrate that CIPER achieves reliable and competitive performance in unified cross-view geo-localization. Benefiting from the orientation-agnostic alignment of the bidirectional cross-attention and the stable spatial reasoning of the set prediction strategy, our method exhibits significantly higher accuracy compared to existing approaches, particularly under challenging conditions involving limited fields of view and arbitrary orientations. As a result, CIPER serves as a robust baseline suitable for diverse real-world applications.

%

% Our contributions are as follows:
% \begin{itemize}
% \item We propose the unified solution to the cross-view image geo-localization problem, providing end-to-end methods for image retrieval and pose estimation.
% \item We propose a cross-view pose estimation decoder based on a two-way transformer, which achieves high accuracy immediately without the need for an iterative estimation process.
% \item Our method significantly outperforms existing methods in pose estimation accuracy for arbitrary orientations, overcoming a major limitation in GPS-denied environments.
% \item Through extensive experiments with diverse large-scale datasets, we consistently demonstrate that the proposed method achieves top-tier performance, proving its suitability for real-world applications.
% \end{itemize}
Our contributions are as follows:
% \begin{itemize}
% \item We redefine cross-view geo-localization as a unified task that simultaneously requires city-scale image retrieval and precise 3-DoF pose estimation, addressing the limitations of treating the two problems separately.
% \item We propose CIPER, a single transformer-based network that jointly performs retrieval and pose estimation, enabling consistent feature learning across a coarse-to-fine localization process within a shared feature space.
% \item We introduce a dual-token transformer encoder to disentangle global retrieval and spatial localization features, along with a two-way transformer–based decoder for robust ground–aerial feature alignment.
% \item We employ a set prediction strategy for reliable, one-shot pose estimation and optimize the entire framework under a unified multi-task objective. Extensive experiments demonstrate strong performance on large-scale benchmarks, effectively overcoming major limitations of visual localization in GPS-denied environments.
% \end{itemize}

\begin{itemize}

\item We redefine cross-view geo-localization as a unified task that simultaneously requires city-scale image retrieval and precise 3-DoF pose estimation, addressing the limitations of treating the two problems separately.

\item We propose CIPER, a unified transformer-based architecture that integrates a dual-token encoder and a two-way decoder, enabling joint learning of retrieval and pose estimation within a shared feature space.

\item We introduce a task-oriented dual-token encoder that learns dedicated tokens for global retrieval and spatial localization while sharing a common visual backbone, allowing flexible feature sharing with task-specific specialization.

% \item Extensive experiments demonstrate that the proposed method achieves strong performance, particularly under large viewpoint uncertainty, while reducing computational redundancy compared to separate retrieval–pose pipelines. These properties make the framework well-suited for real-world applications, where reliable priors are often unavailable and computational efficiency is critical.
\item We demonstrate through extensive experiments that the proposed method achieves strong performance, particularly under large viewpoint uncertainty, while reducing computational redundancy compared to separate retrieval–pose pipelines, making it suitable for real-world applications where reliable priors are often unavailable and computational efficiency is critical.

\end{itemize}

\section{Related Work}

\subsection{Cross-view Image Retrieval}

Cross-view image retrieval involves finding a matching aerial image from a database when given a ground image as a query. The commonly used image retrieval approach involves extracting descriptors from both images and comparing these descriptors to calculate similarity. CVM-Net~\cite{hu2018cvm} proposed a learning-based image retrieval method using a Siamese network~\cite{koch2015siamese} and NetVLAD~\cite{arandjelovic2016netvlad}, Liu et al.~\cite{liu2019lending} proposed enhancing the discriminative power by embedding orientation into the ground and aerial images input to the neural network. L2LTR~\cite{yang2021cross} proposed a self-cross attention mechanism, which has shown improved retrieval performance without increasing the model's complexity.

The difficulty in cross-view image matching is in the significant visual differences between ground and aerial images. To address this issue, an approach has been proposed to reduce the domain gap by transforming images. SAFA~\cite{shi2019spatial} introduced a neural network with a spatial-aware attention module that polar-transforms aerial images into a domain of ground images for comparison. CDTE~\cite{toker2021coming} proposed a method using a generative network to synthesize ground images from aerial images. In IBL~\cite{shi2022accurate}, both polar and projective transforms were utilized to convert aerial images to the domain of ground images. Regmi et al.~\cite{regmi2019bridging} proposed a method using a generative network to synthesize aerial images from queried ground images. In addition, CVFT~\cite{shi2020optimal} proposed a method in which features are extracted from both ground and aerial images, where the ground features are then transformed into aerial-like features to reduce the domain gap for comparison. This approach is intuitive but requires certain assumptions; for example, the viewpoint of the ground image must align with the center of the aerial image. These assumptions limit the applicability of the algorithm.

Recent research has addressed image retrieval under more realistic assumptions, specifically when the viewpoint and center points of two images do not align. VIGOR~\cite{zhu2021vigor} released a large-scale dataset in which the pose of the query ground image exists arbitrarily on the reference aerial image. Additionally, they proposed an image retrieval method in a coarse-to-fine manner, using a Siamese network, the SAFA block, and MLPs. TransGeo~\cite{zhu2022transgeo} proposed a two-stage method using transformers. In the first stage, it employs a transformer encoder to extract features from ground and aerial images. In the second stage, it utilizes an attention-guided cropping method to extract features at a higher resolution for important regions in aerial images. This method has demonstrated state-of-the-art performance.

\subsection{Cross-view Pose Estimation}

Cross-view pose estimation involves estimating a 3-DoF pose, including the offset and orientation between the measured viewpoint of the ground image and the center point of the aerial image. CVML~\cite{xia2022visual} proposed a method that introduces cosine similarity into pose estimation to estimate the dense probability distribution of the ground image on the aerial image. However, this method has limitations, as it relies on the assumption that ground and aerial images are aligned in orientation.

LM~\cite{shi2022beyond} proposed a method capable of 3-DoF pose estimation not only for commonly used 360-degree panoramic ground images but also for ground images with limited fields of view. This method applies geometric projection to aerial image features and iteratively compares with ground image features using LM optimization to estimate the 3-DoF pose. SliceMatch~\cite{lentsch2023slicematch} consists of three stages: feature extraction, aggregation, and final pose estimation. Here, the ``slice" in the aggregation process is designed to handle directional information of the ground image. BoostAcc~\cite{shi2023boosting} aims to enhance performance by decoupling 3-DoF pose estimation. It generates a synthesized overhead view from the query ground image, which is then processed with the aerial image in a neural pose estimator. PureACL~\cite{wang2023pureACL} extracts spatial features from the image to select view-consistent key points and uses these key points to iteratively refine the estimated pose during the optimization process.
%CCVPE~\cite{xia2023convolutional} extracts orientation-aware descriptors from images, performs roll and shift to coarsely estimate the position, and refines the final pose estimation through regression.

%

Thus, the two primary paradigms—image retrieval and pose estimation—for cross-view geo-localization have been briefly discussed. However, for practical deployment, a system must seamlessly integrate a wide search range with high localization accuracy. Rather than cascading two disjoint models which leads to redundant feature extraction and computational inefficiency, we advocate for a unified approach. In this work, we propose an end-to-end framework where coarse retrieval and fine-grained 3-DoF pose estimation are jointly learned and executed within a shared feature space, ensuring robust performance from city-scale localization down to precise alignment.

\section{Methods}

\subsection{Problem Statement}

The cross-view image geo-localization problem is defined as follows. Given a ground image query and a database of aerial images, the first task is image retrieval, which finds the aerial image containing the location of the ground image. The second task is pose estimation, which estimates the precise 3-DoF position of the ground image on the aerial image.

\subsection{Network Structure}

The proposed network, CIPER, is a single network with two submodules that enables end-to-end learning and inference of image retrieval and pose estimation. The overall structure of CIPER is illustrated in Fig.~\ref{fig:mtd_overivew}.

\subsubsection{Image Encoder}

To jointly address image retrieval and pose estimation, the encoder must capture two complementary representations: a compact and discriminative global descriptor for large-scale retrieval, and spatially-sensitive local features that preserve geometric structure for precise pose estimation. Retrieval relies on holistic semantic understanding of the scene, whereas pose estimation requires fine-grained spatial cues. Therefore, a unified backbone must simultaneously encode global context and spatial structure.

Convolutional neural networks gradually enlarge receptive fields through hierarchical stacking but remain biased toward local interactions. In contrast, Vision Transformers (ViT)~\cite{dosovitskiy2020image} enable direct global interactions among all image patches via full self-attention at every layer, which is particularly advantageous for cross-view geo-localization where correspondences between ground and aerial views are often spatially non-local and geometrically distorted. We adopt the original ViT formulation rather than hierarchical or window-based variants, as its full global attention facilitates early cross-view alignment under large viewpoint gaps. Moreover, its flat token structure naturally supports the integration of task-specific tokens, such as the class and pose tokens in our unified retrieval–pose framework, enabling consistent multi-task feature learning within a single backbone.

Given an input image $I \in \mathbb{R}^{h \times w \times c}$, we first divide it into non-overlapping patches of size $ps \times ps$. This is implemented using a Conv2D layer with kernel size $ps \times ps$, stride $ps$, and output dimension $d$. The operation linearly projects each image patch into a $d$-dimensional embedding vector. After reshaping and normalization, the input image is transformed into a sequence of patch embeddings $PE \in \mathbb{R}^{h'w' \times d}$, where $h' = \frac{h}{ps}$ and $w' = \frac{w}{ps}$. Each row of $PE$ corresponds to a patch-level representation, preserving spatial granularity while enabling global attention.

To disentangle retrieval and localization objectives within a shared backbone, we introduce two learnable tokens: a class token and a pose token. These tokens are initialized as trainable embeddings and concatenated with the patch embeddings, $X_0 = [cls\_token,\, pos\_token,\, PE]$.
The class token is designed to aggregate global contextual information through self-attention and serves as a compact descriptor for image retrieval. In contrast, the pose token interacts with spatial patch embeddings while retaining localization-sensitive information for downstream pose estimation.

The combined token sequence $X_0$ is processed by a multi-layer transformer encoder consisting of 12 layers with 6 attention heads and learnable positional embeddings. The embedding dimension is set to $d = 384$, and the patch size $ps$ is set to 16 or 32. 

Through multi-head self-attention, all tokens—including the class and pose tokens—can attend to the entire patch sequence, enabling long-range interactions and global context modeling. The encoder outputs three components:
\begin{itemize}
    \item $cls\_token$: a global descriptor used for image retrieval,
    \item $pos\_token$: a localization-aware token used for pose decoding,
    \item $img\_embed \in \mathbb{R}^{h'w' \times d}$: spatial embeddings used for cross-view alignment.
\end{itemize}

This design allows a single transformer backbone to produce task-specialized representations while maintaining computational efficiency. The class token captures discriminative global semantics for large-scale retrieval, whereas the pose token and spatial embeddings preserve geometric structure necessary for precise 3-DoF pose estimation.

\subsubsection{Pose Decoder}

\begin{figure*}[t]
    \begin{center}
    \includegraphics[width=0.6\linewidth]{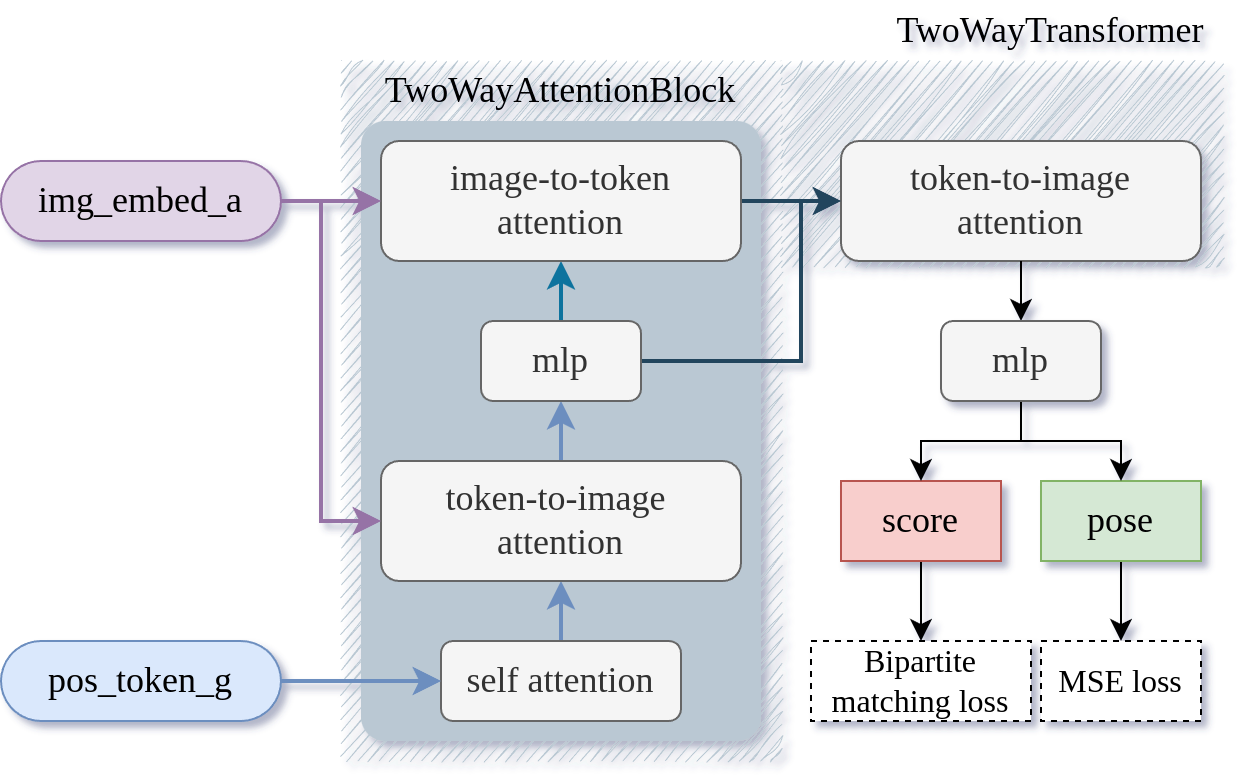}
    \end{center}
    \vspace*{-0.35cm}
    \caption{\textbf{Structure of the pose decoder} 
    The pose decoder is composed of a two-way transformer with a two-way attention block. The decoder takes \textit{pos\_token\_g} as a prompt token and \textit{img\_embed\_a} as an image embedding, producing 3-DoF pose as the output.
    }
    \label{fig:mtd_decoder}
\end{figure*}

% In cross-view pose estimation, the most challenging problem is the large domain gap between ground and aerial images. To address this issue, the mask decoder structure from Segment Anything \cite{kirillov2023segment} is adopted into our design as a two-way transformer. The two-way transformer is initially proposed for prompt segmentation, which imports a linguistic prompt to generate segmentation masks from images. It successfully generates segmentation results highly relevant to the given prompts by employing bidirectional attention from token-to-image and image-to-token.

% In the proposed pose decoder, the two-way transformer uses attention mechanisms both from ground to aerial and from aerial to ground. This enables the prediction of a highly relevant 3-DoF pose in the aerial image based on the provided ground image prompt. To this end, the ground image feature $(pos\_token\_g \in \mathbb{R}^{1 \times 1 \times d})$ and the aerial image feature $(img\_embed\_a \in \mathbb{R}^{1 \times h’w’ \times d})$ are used as the prompt token and image embedding, respectively. 

% Finally, the set prediction method proposed in DETR~\cite{carion2020end} is employed for robust pose estimation. The prompt token is input into a two-way transformer as queries, and the output is passed to an MLP. The MLP predicts the score and 3-DoF pose for each of the $q$ queries. The query with the highest score becomes the final pose estimation result.  The number of queries $q$ used was 64. The structure of the pose decoder is shown in Fig.~\ref{fig:mtd_decoder}.

In cross-view pose estimation, the primary challenge lies in the substantial domain gap between ground and aerial images. The two views exhibit significant differences in viewpoint, scale, orientation, and geometric structure, making direct feature matching unreliable. To bridge this gap, we design a two-way transformer for cross-view pose estimation, taking inspiration from the highly effective bidirectional attention mechanism introduced in Segment Anything~\cite{kirillov2023segment}.

We adapt this architecture to enable deep reciprocal interaction between spatial queries and image embeddings through alternating token-to-image and image-to-token cross-attention. This bidirectional mechanism allows the network to mutually refine representations in a single forward pass by conditioning aerial features on ground-based queries and vice versa. We leverage this property to align heterogeneous ground and aerial features effectively without relying on complex iterative optimization procedures.

In our pose decoder, the ground image pose token 
\(
pos\_token\_g \in \mathbb{R}^{1 \times 1 \times d}
\)
is used as a spatial query token, while the aerial image embedding
\(
img\_embed\_a \in \mathbb{R}^{1 \times h'w' \times d}
\)
serves as the spatial memory. Through cross-attention from ground to aerial features, the pose token aggregates spatially relevant information from the aerial image. Conversely, aerial features are refined by attending to the ground query, enabling robust alignment between the two domains. This reciprocal bidirectional interaction effectively mitigates viewpoint discrepancies and enhances cross-view correspondence modeling.

To further improve robustness in pose regression, we adopt the set prediction paradigm introduced in DETR~\cite{carion2020end}. Instead of predicting a single pose hypothesis, we employ $q$ learnable queries ($q=64$) that are processed by the two-way transformer. Each query produces a candidate 3-DoF pose along with a confidence score through a lightweight multi-layer perceptron (MLP). The final pose estimate is selected based on the highest confidence score. This set-based formulation improves stability under ambiguous cross-view matches and reduces sensitivity to noisy alignments. The overall structure of the proposed pose decoder is illustrated in Fig.~\ref{fig:mtd_decoder}.

\subsection{Loss Function}

The loss functions are as follows:
\begin{equation}
\begin{aligned}
    &\mathcal{L}_{ir} = \log(1+e^{\alpha(d_{pos}-d_{neg})}) \\
    &\mathcal{L}_{pe} = \lambda_{cls} L_{bce} + L_{mse} 
     \label{eqn:loss}
\end{aligned}
\end{equation}

First, the triplet loss, denoted as $\mathcal{L}_{ir}$, is used for image retrieval. The triplet loss increases the similarity between positive pairs $d_{pos}$ and decreases the similarity between negative pairs $d_{neg}$. 
Similarity is calculated as the result of a matrix multiplication between class tokens.

Second, the binary cross-entropy (BCE) loss and MSE loss, denoted as $\mathcal{L}_{pe}$, are used for pose estimation. BCE loss is computed for pairs generated from the bipartite matching results in set prediction. The network learns to estimate the query that best matches the target among $q$ queries through BCE loss. In addition, MSE loss is employed to regress the 3-DoF pose value from the queries. $\lambda_{cls}$ was set to 0.2.

\section{Experiments}
\subsection{Experimental Settings}

\subsubsection{Datasets}

VIGOR~\cite{zhu2021vigor} dataset is a large-scale cross-view image dataset, where the viewpoint of the ground image exists at an arbitrary location within the aerial image. The experiment utilized 105,214 ground panorama images ($\pm 180^\circ$ field of view) and 90,618 aerial images collected from four U.S. cities. The positions of positive samples have an IoU (Intersection over Union) greater than 0.39 with the center of the aerial image. VIGOR provides two train / validation splits. The ``same area” split uses data from all four cities for both training and validation. The ``cross area” split uses data from two cities (NewYork, Seattle) for training, and data from the remaining two cities (SanFrancisco, Chicago) for validation.

The KITTI~\cite{geiger2013vision} dataset provides image data collected from cameras mounted on vehicles and GPS data. LM~\cite{shi2022beyond} collected an aerial image database based on the GPS data of KITTI. We used ground images from KITTI RGB image data and aerial images from LM. The data were split into training (19,655 samples), test1 (3,773 samples), and test2 (7,542 samples). The data for training and test1 were collected in the same area, test2 in a different area. The ground images have a $\pm 47^\circ$ field of view.

The Ford multi-AV~\cite{agarwal2020ford} dataset provides ground images collected from cameras mounted on vehicles along with GPS and calibration data. LM collected an aerial image database based on data from this dataset. The experiments used two sets (Log1 and Log2). Log1 consists of 4,000 training and 2,100 testing samples, and Log2 consists of 10,350 training and 3,727 testing samples. The ground images have a $\pm 40^\circ$ field of view.

\begin{table*}[t]
\centering
\caption{\textbf{Computational cost comparison between the conventional two-stage pipeline and the proposed simultaneous framework.} }
\label{tab:efficiency}
\vspace*{-0.35cm}
\begin{adjustbox}{width=0.45\linewidth}
\begin{tabular}{|c|c|cc|}

% \begin{tabular}{lccc}
\hline
Task & Method & FLOPs (G) & Params (M) \\
\hline \hline
IR & SAFA~\cite{shi2019spatial} & 84.63& 29.53\\
IR & VIGOR~\cite{zhu2021vigor} & 188.11& 34.02\\
IR & CVML~\cite{xia2022visual} & 330.21& 79.69\\
\hline
PE & LM~\cite{shi2022beyond} & 316.30 & 5.04 \\
% PE & SliceMatch~\cite{lentsch2023slicematch} & XX.X & XX.X \\ % no code
PE & BoostAcc~\cite{shi2023boosting} & 387.36 & 7.22 \\
PE & PureACL~\cite{wang2023pureACL} &  1576.50& 15.71\\
% \hline
% IR + PE & Retrieval + Pose & XX.X & XX.X \\
\hline
IR + PE & \textbf{Ours}& \textbf{24.28} & \textbf{52.54} \\
\hline
% \end{tabular}
\end{tabular}
\end{adjustbox}
\end{table*}
\subsubsection{Implementation Details}

The image resolutions were as follows: $256 \times 1024$ for ground images and $512 \times 512$ for aerial images. The batch size was set to 12. The optimizer used was AdamW, with a learning rate and weight decay of 0.0001. The patch size $ps$ was set to 16 for VIGOR and 32 for KITTI and Ford multi-AV.

\subsection{Computational Efficiency Analysis}

Conventional cross-view geo-localization pipelines typically treat image retrieval and pose estimation as two separate stages. 
%A retrieval network first searches a large aerial database to identify candidate locations, and a separate pose estimation network then refines the relative translation and orientation between the ground and aerial images. 
This two-stage design requires both models to independently process the same visual inputs, resulting in redundant feature extraction and increased computational cost.

To quantify this inefficiency, we compare the computational cost of our method with that of image retrieval (IR) and pose estimation (PE) methods. As shown in Tab.~\ref{tab:efficiency}, the conventional pipeline requires the sum of the computational costs of the two networks, whereas the proposed method performs both tasks simultaneously within a single transformer-based architecture. By sharing a unified backbone and intermediate representations, our model avoids redundant computation and achieves both retrieval and pose estimation with fewer FLOPs than the combined cost of the two separate methods, demonstrating the efficiency of the proposed framework.

\subsection{Visualization of Unified Cross-view Geo-localization}

\begin{figure*}[t]
    \begin{center}
    \includegraphics[width=0.7\linewidth]{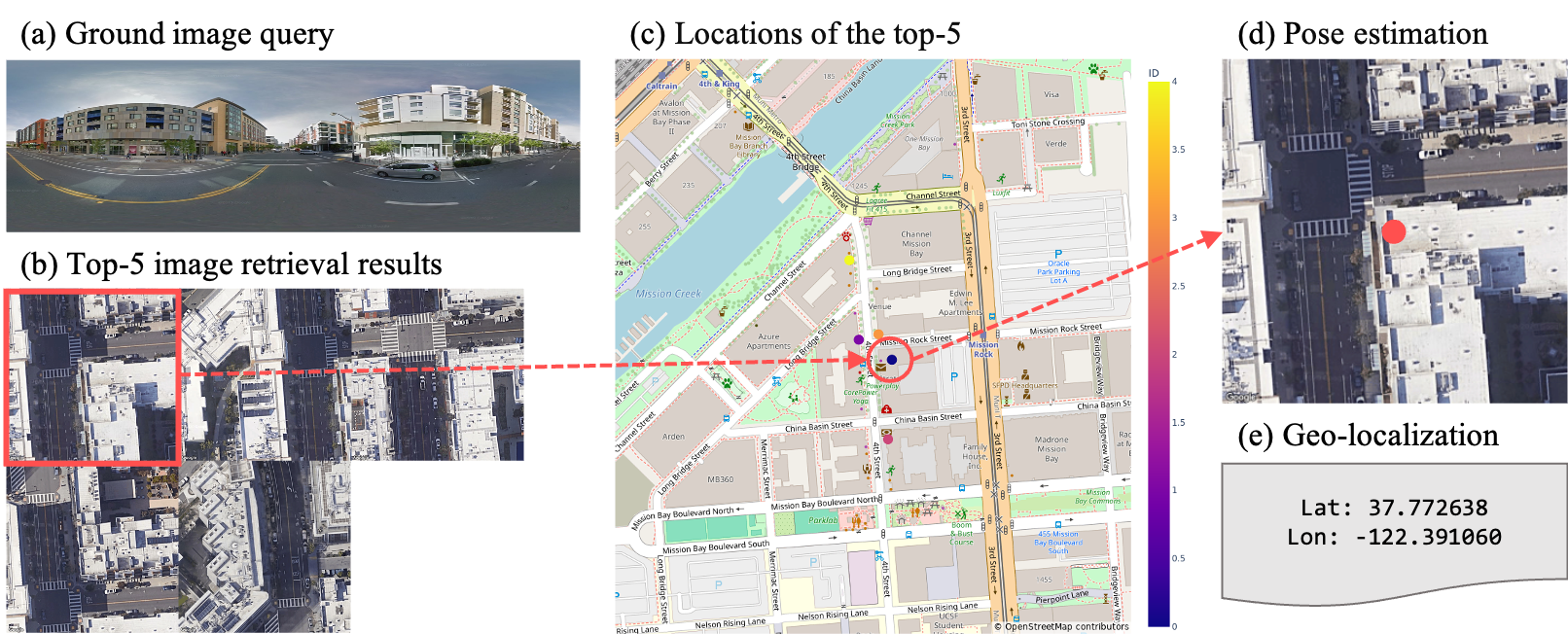}
    \end{center}
    \vspace*{-0.5cm}
    \caption{\textbf{Visualization of cross-view geo-localization results on VIGOR Dataset}
    Fig. 4a shows a ground image query, and Fig. 4b presents the top 5 aerial image candidates from the image retrieval results. In Fig. 4c, the map displays the locations of these top 5 candidates, with colors closer to indigo, indicating a higher similarity score. Fig. 4d illustrates the estimated pose from the aerial image with the highest similarity score. Finally, Fig. 4e presents the final geo-localization result, obtained by combining the location from the image retrieval result with the pose estimation result.}
    \label{fig:exp_vigor}
\end{figure*}

The process and results of cross-view image geo-localization on VIGOR is illustrated in Fig.~\ref{fig:exp_vigor}. Fig.~\ref{fig:exp_vigor}a shows the ground image query, and Fig.~\ref{fig:exp_vigor}b presents the top-5 aerial image candidates selected from the aerial image database as a result of the image retrieval. The locations of the candidates are shown in Fig.~\ref{fig:exp_vigor}c. Among them, pose estimation is performed on the image with the highest similarity, as shown in Fig.~\ref{fig:exp_vigor}d. The final geo-localization result is predicted by combining both the image retrieval and pose estimation results in Fig.~\ref{fig:exp_vigor}e.

In Fig.~\ref{fig:exp_vigor}c, it is noteworthy that the top-5 candidates plotted on the map are closely located. This suggests that precise geo-localization results can be achieved not only when performing pose estimation for the top-1 candidate but also for the top-5 candidates. The experiments using VIGOR were conducted under the assumption that the orientations of the ground and aerial images are aligned, so the results are expressed only in terms of latitude and longitude.

\begin{table*}[t!]
\caption{\textbf{Comparison on cross-view image retrieval and pose estimation using VIGOR dataset} \color{red}Red \color{black}and \color{blue}blue \color{black}represent the best and second-best performances, respectively. Results marked with $\star$ are sourced
directly from the respective original papers.}
\vspace*{-0.35cm}
\centering
\begin{adjustbox}{width=0.6\linewidth}
\begin{tabular}{|c|c|cccc|cc|}
\hline 
\multirow{2}{*}{Area} & \multirow{2}{*}{Method} & \multicolumn{4}{c|}{$(\uparrow)$ Image retrieval $(\%)$} & \multicolumn{2}{c|}{$(\downarrow)$ Pose estimation $(m)$}                                \\ \cline{3-8} 
                       % &                       & \multicolumn{4}{c|}{}                                 & \multicolumn{2}{c|}{}      \\ \cline{3-8} 
                       &                       & $R@1$         & $R@5$        & $R@10$        & $R@1\%$        & Mean & \multicolumn{1}{c|}{Median}  \\ \hline \hline
\multirow{5}{*}{same} & $\star$SAFA~\cite{shi2019spatial} & 33.93 & 58.42 & 68.12 & 98.24 & - & \multicolumn{1}{c|}{-} \\ 
& $\star$VIGOR~\cite{zhu2021vigor} & \color{red}\textbf{41.07} & \color{blue}\textbf{65.81} & \color{blue}\textbf{74.05} & \color{blue}\textbf{98.37} & - & \multicolumn{1}{c|}{-} \\ 
& $\star$CVML~\cite{xia2022visual} & - & - & - & - & 6.94 & \multicolumn{1}{c|}{\color{blue}\textbf{3.64}}\\ 
& $\star$SliceMatch~\cite{lentsch2023slicematch} & - & - & - & - & \color{red}\textbf{5.18} & \multicolumn{1}{c|}{\color{red}\textbf{2.58}}\\ 
% BoostAcc~\cite{shi2023boosting} & same & - & - & - & - & 4.12 & \multicolumn{1}{c|}{1.34} \\ 
& Ours & \color{blue}\textbf{40.98} & \color{red}\textbf{67.55} & \color{red}\textbf{75.94} & \color{red}\textbf{98.67} & \color{blue}\textbf{5.25} & \multicolumn{1}{c|}{4.11} \\ \hline \hline

\multirow{5}{*}{cross} & $\star$SAFA~\cite{shi2019spatial} & 8.2 & 19.59 & 26.36 & 77.61 & - & \multicolumn{1}{c|}{-} \\ 
& $\star$VIGOR~\cite{zhu2021vigor}  & \color{red}\textbf{11} & \color{blue}\textbf{23.56} & \color{blue}\textbf{30.76} & \color{blue}\textbf{80.22} & - & \multicolumn{1}{c|}{-} \\ 
& $\star$CVML~\cite{xia2022visual} & - & - & - & - & 9.05 & \multicolumn{1}{c|}{5.14}\\ 
& $\star$SliceMatch~\cite{lentsch2023slicematch} & - & - & - & - & \color{red}\textbf{5.53} & \multicolumn{1}{c|}{\color{red}\textbf{2.55}} \\ 
% BoostAcc~\cite{shi2023boosting} & cross & - & - & - & - & 5.16 & \multicolumn{1}{c|}{1.4}  \\ 
& Ours & \color{blue}\textbf{10.53} & \color{red}\textbf{24.05} & \color{red}\textbf{31.92} & \color{red}\textbf{82.44} & \color{blue}\textbf{6.2} & \multicolumn{1}{c|}{\color{blue}\textbf{4.79}}\\ \hline

\end{tabular}
\end{adjustbox}

\label{tab:exp_vigor}
\end{table*}

\begin{table*}[t!]
\caption{\textbf{Comparison on cross-view pose estimation using KITTI dataset} \color{red}Red \color{black}and \color{blue}blue \color{black}represent the best and second-best performances, respectively. Results marked with $\star$ are sourced
directly from the respective original papers.}
\vspace*{-0.35cm}
\centering
\begin{adjustbox}{width=1\linewidth}
\begin{tabular}{|c|c|c|cc|cc|cc|cc|cc|}
\hline
 &  &  & \multicolumn{2}{c|}{$(\downarrow)$ Location $(m)$} & \multicolumn{2}{c|}{$(\uparrow)$ Latitude $(\%)$} & \multicolumn{2}{c|}{$(\uparrow)$ Longitude $(\%)$} & \multicolumn{2}{c|}{$(\downarrow)$ Orientation $(^\circ)$} & \multicolumn{2}{c|}{$(\uparrow)$ Orientation $(\%)$} \\ \cline{4-13} 
 Area & Prior & Method & Mean & Median & $R@1m$ & $R@5m$ & $R@1m$ & $R@5m$ & Mean & Median & $R@1^\circ$ & $R@5^\circ$ \\ \hline \hline

\multirow{5}{*}{same} & \multirow{5}{*}{$\pm 10^\circ$} & LM~\cite{shi2022beyond} 
 & 12.08 & 11.44 & 33.74 & 79.57 & 6.25 & 27.06 & 3.72 & 2.83 & 20.67 & 72.41 \\
& & $\star$SliceMatch~\cite{lentsch2023slicematch} 
 & 7.96 & 4.39 & 49.09 & 98.52 & 15.19 & 57.35 & 4.12 & 3.65 & 13.41 & 64.17 \\
& & BoostAcc~\cite{shi2023boosting}
 & 7.87 & 3.17 & 76.41 & \color{blue}\textbf{98.89} & 23.51 & 62.23 & \color{red}\textbf{0.28} & \color{red}\textbf{0.23} & \color{red}\textbf{99.07} & \color{red}\textbf{100} \\
& & PureACL~\cite{wang2023pureACL} & \color{blue}\textbf{2.42} & \color{red}\textbf{0.42} & \color{red}\textbf{91.95} & 94.28 & \color{red}\textbf{91.86} & \color{blue}\textbf{92.38}& 3.97& 1.77& 32.71&71.66\\
& & Ours 
 & \color{red}\textbf{2.02} & \color{blue}\textbf{1.38} & \color{blue}\textbf{88.74} & \color{red}\textbf{99.81} & \color{blue}\textbf{42.35} & \color{red}\textbf{94.04} & \color{blue}\textbf{0.69} & \color{blue}\textbf{0.44} & \color{blue}\textbf{80.28} & \color{blue}\textbf{99.55} \\ \hline

\multirow{5}{*}{same} & \multirow{5}{*}{$\pm 180^\circ$}& LM~\cite{shi2022beyond} 
 & 14.92 & 15.46 & 4 & 18 & 10 & 35 & 94.72 & 94.35 & 0 & 3 \\
& & $\star$SliceMatch~\cite{lentsch2023slicematch} 
 & \color{blue}\textbf{9.39} & \color{red}\textbf{5.41} & \color{blue}\textbf{39.73} & \color{blue}\textbf{87.92} & \color{blue}\textbf{13.63} & \color{blue}\textbf{49.22} & \color{blue}\textbf{8.71} & \color{blue}\textbf{4.42} & \color{blue}\textbf{11.35} & \color{blue}\textbf{55.82} \\
& & BoostAcc~\cite{shi2023boosting}
 & 19.39 & 17.72 & 8.43 & 41.21 & 5.43 & 23.22 & 90.02 & 90.24 & 0.56 & 2.73 \\
& & PureACL~\cite{wang2023pureACL} & 13.86& 12.31& 10.26& 33.41& 8.29& 25.52& 90.04& 90.10& 1.34&4.20\\
& & Ours 
 & \color{red}\textbf{8.26} & \color{blue}\textbf{6.81} & \color{red}\textbf{64.91} & \color{red}\textbf{89.85} & \color{red}\textbf{55.13} & \color{red}\textbf{74.48} & \color{red}\textbf{3.29} & \color{red}\textbf{1.31} & \color{red}\textbf{40.13} & \color{red}\textbf{90.19} \\ \hline \hline 
\end{tabular}
\end{adjustbox}

\begin{adjustbox}{width=1\linewidth}
\begin{tabular}{|c|c|c|cc|cc|cc|cc|cc|}
\hline
 &  &  & \multicolumn{2}{c|}{$(\downarrow)$ Location $(m)$} & \multicolumn{2}{c|}{$(\uparrow)$ Latitude $(\%)$} & \multicolumn{2}{c|}{$(\uparrow)$ Longitude $(\%)$} & \multicolumn{2}{c|}{$(\downarrow)$ Orientation $(^\circ)$} & \multicolumn{2}{c|}{$(\uparrow)$ Orientation $(\%)$} \\ \cline{4-13} 
 Area & Prior & Method & Mean & Median & $R@1m$ & $R@5m$ & $R@1m$ & $R@5m$ & Mean & Median & $R@1^\circ$ & $R@5^\circ$ \\ \hline \hline
\multirow{5}{*}{cross} & \multirow{5}{*}{$\pm 10^\circ$} & LM~\cite{shi2022beyond} 
 & 12.58 & 12.12 & 26.96 & 72.82 & 5.2 & 26.45 & 3.95 & 3.03 & 18.52 & 70.51 \\
& & $\star$SliceMatch~\cite{lentsch2023slicematch} 
 & 13.5 & 9.77 & 32.43 & 86.44 & 8.3 & 35.57 & 4.2 & 6.61 & 46.82 & 46.82 \\
& & BoostAcc~\cite{shi2023boosting}
 & 11.31 & 6.83 & \color{blue}\textbf{57.74} & \color{blue}\textbf{91.15} & \color{blue}\textbf{14.16} & 44.97 & \color{red}\textbf{0.28} & \color{red}\textbf{0.23} & \color{red}\textbf{98.98} & \color{red}\textbf{100} \\
& & PureACL~\cite{wang2023pureACL} & \color{red}\textbf{6.20}& \color{red}\textbf{0.61}& \color{red}\textbf{67.24}& \color{red}\textbf{92.76}& \color{red}\textbf{64.81}& \color{red}\textbf{89.69}& 4.26& 2.48& 23.45&59.24\\
& & Ours 
 & \color{blue}\textbf{9.28} & \color{blue}\textbf{6.44} & 39.45 & 88.4 & 12.34 & \color{blue}\textbf{46.82} & \color{blue}\textbf{1.76} & \color{blue}\textbf{1.06} & \color{blue}\textbf{47.76} & \color{blue}\textbf{93.04} \\ \hline

\multirow{5}{*}{cross} & \multirow{5}{*}{$\pm 180^\circ$}& LM~\cite{shi2022beyond} 
 & {14.74} & 15.44 & 1 & 15 & 10 & 36 & 94.69 & 97.06 & 0 & 2 \\
& & $\star$SliceMatch~\cite{lentsch2023slicematch} 
 & \color{blue}\textbf{14.85} & \color{blue}\textbf{11.85} & \color{blue}\textbf{24} & \color{blue}\textbf{72.89} & 7.17 & 33.12 & \color{red}\textbf{23.64} & \color{blue}\textbf{7.96} & \color{red}\textbf{31.69} & \color{blue}\textbf{31.69} \\
& & BoostAcc~\cite{shi2023boosting}
 & 20.48 & 19.17 & 8.71 & 39.25 & 5.60 & 21.44 & 89.85 & 89.86 & 0.53 & 2.85 \\
& & PureACL~\cite{wang2023pureACL} & 15.14& 14.74& 9.70& 41.97& \color{blue}\textbf{9.75}& \color{blue}\textbf{35.73}& 90.05& 90.19& 0.97&2.53\\
& & Ours 
 & \color{red}\textbf{12.54} & \color{red}\textbf{11.33} & \color{red}\textbf{56.93} & \color{red}\textbf{77.13} & \color{red}\textbf{53.47} & \color{red}\textbf{67.24} & \color{blue}\textbf{25.18} & \color{red}\textbf{4.48} & \color{blue}\textbf{15.45} & \color{red}\textbf{52.64} \\ \hline
\end{tabular}
\end{adjustbox}

\label{tab:exp_kitti}
\end{table*}

\begin{table*}[t!]
\caption{\textbf{Comparison on cross-view pose estimation using Ford multi-AV dataset} \color{red}Red \color{black}and \color{blue}blue \color{black}represent the best and second-best performances, respectively.}
\centering
\vspace*{-0.35cm}
\begin{adjustbox}{width=1\linewidth}
\begin{tabular}{|c|c|c|cc|cc|cc|cc|cc|}

\hline
 &  &  & \multicolumn{2}{c|}{$(\downarrow)$ Location $(m)$} & \multicolumn{2}{c|}{$(\uparrow)$ Latitude $(\%)$} & \multicolumn{2}{c|}{$(\uparrow)$ Longitude $(\%)$} & \multicolumn{2}{c|}{$(\downarrow)$ Orientation $(^\circ)$} & \multicolumn{2}{c|}{$(\uparrow)$ Orientation $(\%)$} \\ \cline{4-13} 
 Area & Prior & Method & Mean & Median & $R@1m$ & $R@5m$ & $R@1m$ & $R@5m$ & Mean & Median & $R@1^\circ$ & $R@5^\circ$ \\ \hline \hline

\multirow{3}{*}{Log1} & \multirow{3}{*}{$\pm 10^\circ$} & LM~\cite{shi2022beyond} 
 & \color{blue}\textbf{12.54} &  \color{blue}\textbf{12.65} & \color{blue} \textbf{48.57} & 71.57 & \color{blue} \textbf{5.86} & \color{blue} \textbf{26.33} & 3.13 & 1.29 & 42.86 & 79.62 \\
& & BoostAcc~\cite{shi2023boosting}
& 16.06 & 14.96 & \color{red} \textbf{67.57} &  \color{red}\textbf{95.95} &  \color{red}\textbf{18.19} &  \color{red}\textbf{29.24} &  \color{blue}\textbf{1.27} &  \color{blue}\textbf{0.65} &  \color{blue}\textbf{68.43} & \color{red}\textbf{95.05} \\
& & Ours 
& \color{red}\textbf{10.21} & \color{red}\textbf{10.14} & 48.24 & \color{blue}\textbf{90.14} & 5.24 & 25.43 & \color{red}\textbf{1.21} & \color{red}\textbf{0.63} & \color{red}\textbf{69.81} &  \color{red}\textbf{95.05} \\ \hline

\multirow{3}{*}{Log1} & \multirow{3}{*}{$\pm 180^\circ$} & LM~\cite{shi2022beyond} 
 &  \color{blue}\textbf{15.44} &  \color{blue}\textbf{16.08} & 4.95 & 26.38 & 5.10 & \color{blue}\textbf{23.29} & 90.23 & 90.81 & 0.62 & 2.80 \\
& & BoostAcc~\cite{shi2023boosting}
 & 23.05 & 20.91 &  \color{blue}\textbf{6.19} &  \color{blue}\textbf{30.43} &  \color{blue}\textbf{5.71} &  {20.43} &  \color{blue}\textbf{47.16} &  \color{blue}\textbf{34.56} &  \color{blue}\textbf{1.10} &  \color{blue}\textbf{7.05} \\
& & Ours 
 & \color{red}\textbf{10.73} & \color{red}\textbf{9.96} & \color{red}\textbf{60.95} & \color{red}\textbf{83.52} & \color{red}\textbf{53} & \color{red}\textbf{68.38} & \color{red}\textbf{4.41} & \color{red}\textbf{0.62} & \color{red}\textbf{68.62} & \color{red}\textbf{90.43} \\ \hline \hline

\end{tabular}
\end{adjustbox}

\begin{adjustbox}{width=1\linewidth}
\centering
\begin{tabular}{|c|c|c|cc|cc|cc|cc|cc|}

\hline
 &  &  & \multicolumn{2}{c|}{$(\downarrow)$ Location $(m)$} & \multicolumn{2}{c|}{$(\uparrow)$ Latitude $(\%)$} & \multicolumn{2}{c|}{$(\uparrow)$ Longitude $(\%)$} & \multicolumn{2}{c|}{$(\downarrow)$ Orientation $(^\circ)$} & \multicolumn{2}{c|}{$(\uparrow)$ Orientation $(\%)$} \\ \cline{4-13} 
 Area & Prior & Method & Mean & Median & $R@1m$ & $R@5m$ & $R@1m$ & $R@5m$ & Mean & Median & $R@1^\circ$ & $R@5^\circ$ \\ \hline \hline
\multirow{3}{*}{Log2} & \multirow{3}{*}{$\pm 10^\circ$} &LM~\cite{shi2022beyond} 
 &  \color{blue}\textbf{12.01} &  \color{blue}\textbf{11.50} & \color{blue} \textbf{29.97} & 77.78 & 4.97 & 26 & 4.36 & 3.75 & 15.29 & 63.03 \\
& & BoostAcc~\cite{shi2023boosting}
 & 15.77 & 14.58 &  \color{red}\textbf{67.96} &  \color{red}\textbf{96.16} &  \color{red}\textbf{14.92} &  \color{red}\textbf{31.71} & \color{blue}\textbf{0.88} &  \color{blue}\textbf{0.60} &  \color{blue}\textbf{69.55} & \color{blue}\textbf{99.03} \\
& & Ours 
 & \color{red}\textbf{11.19} & \color{red}\textbf{9.75} & 22.73 & \color{blue}\textbf{85.19} & \color{blue}\textbf{6.36} & \color{blue}\textbf{29.33} &  \color{red}\textbf{0.66} & \color{red}\textbf{0.38} & \color{red}\textbf{84.33} &  \color{red}\textbf{99.09} \\ \hline

\multirow{3}{*}{Log2} & \multirow{3}{*}{$\pm 180^\circ$} & LM~\cite{shi2022beyond} 
 &  \color{blue}\textbf{15.71} &  \color{blue}\textbf{16.08} & 4.86 & 25.36 & 4.80 & \color{blue}\textbf{24.60} & 90.03 & 90.59 & 0.56 & 2.66 \\
& & BoostAcc~\cite{shi2023boosting}
 & 22.28 & 20.08 &  \color{blue}\textbf{4.91} &  \color{blue}\textbf{26.03} &  \color{blue}\textbf{6.47} &  {21.55} &  \color{blue}\textbf{53.25} &  \color{blue}\textbf{42.49} &  \color{blue}\textbf{1.40} &  \color{blue}\textbf{6.25} \\
& & Ours 
 & \color{red}\textbf{11.83} & \color{red}\textbf{10.92} & \color{red}\textbf{57.69} & \color{red}\textbf{80.25} & \color{red}\textbf{51.49} & \color{red}\textbf{66.03} & \color{red}\textbf{9.49} & \color{red}\textbf{1.02} & \color{red}\textbf{48.83} & \color{red}\textbf{89.51} \\ \hline
\end{tabular}
\end{adjustbox}

\label{tab:exp_ford}
\end{table*}

\subsection{Experiments on Cross-view Image Retrieval}

We compared the accuracies of the cross-view image retrieval methods on the VIGOR dataset using top-k recall accuracy ($r@k$) as the evaluation metric. The top-k predictions for each test query were defined as the k reference images with the highest similarity scores. A prediction was considered correct if a ground-truth image was included in the top-k predictions. The experimental results are presented in Tab.~\ref{tab:exp_vigor}. 

Our method outperformed other state-of-the-art image retrieval methods in terms of $r@5$,  $r@10$ and  $r@1\%$ metrics. Additionally, it showed comparable performance to other methods at the $r@1$. This result shows the validity of our method in image retrieval. It accurately searches candidates from a city-scale aerial image database, ensuring high accuracy in geo-localization.

\subsection{Experiments on Cross-view Pose Estimation}

\begin{figure*}[t]
    \begin{center}
    \includegraphics[width=1\linewidth]{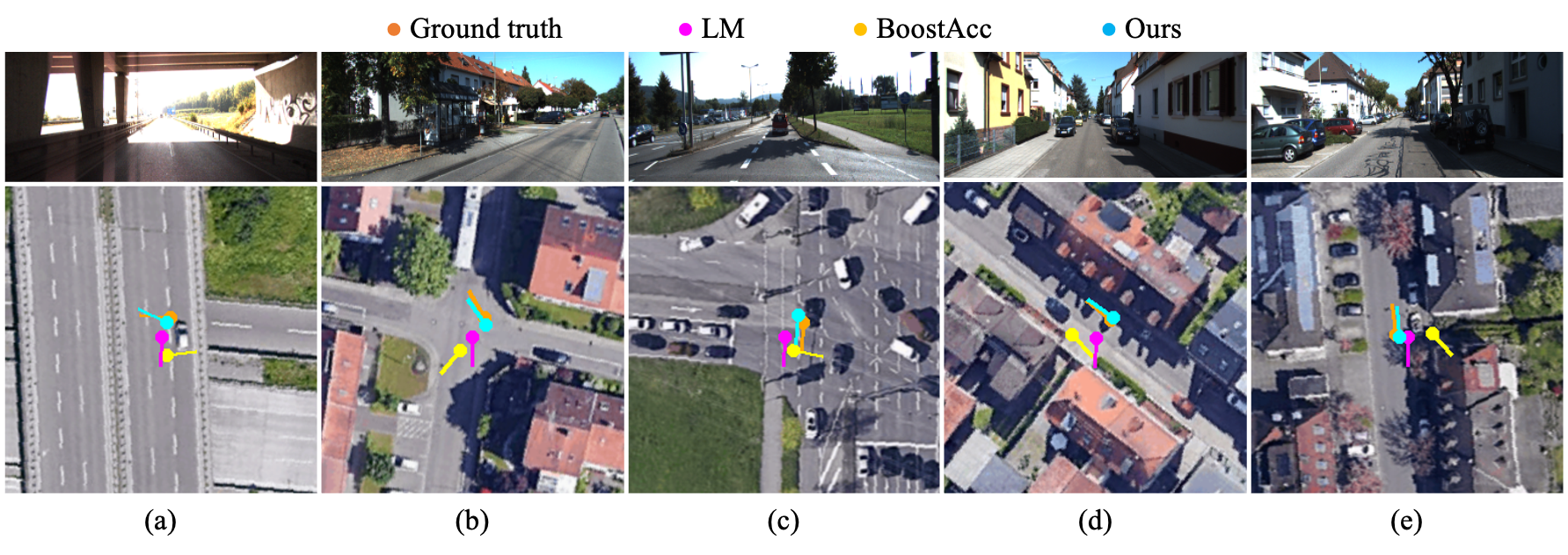}
    \end{center}
    \vspace*{-0.5cm}
    \caption{\textbf{Visualization of cross-view pose estimation results on the KITTI dataset} 
    The first row shows the ground image query and the second row shows the pose estimation results on the aerial image. The orange, magenta, yellow and cyan marks indicate the ground truth, LM, BoostAcc and Our prediction, respectively. 
    }
    \label{fig:exp_pe_kitti}
\end{figure*}

In this section, we compared the accuracies of the cross-view image pose estimation methods. The experimental results are presented in Tab.~\ref{tab:exp_kitti} and Tab.~\ref{tab:exp_ford} for the KITTI and Ford Multi-AV datasets, respectively. Here, \textit{prior} denotes the rotation range of aerial images. The aerial images were randomly rotated within the specified \textit{prior} range. In addition, a random translation in the range of $\pm 20 m$ was applied in all experiments.

%

% The evaluation metric included the mean and median values for the differences in location $(x, y)$ and orientation $(\theta)$ between the ground truth and prediction. Smaller values indicated better performance. We also used the evaluation metric presented in SliceMatch~\cite{lentsch2023slicematch}. This metric includes recall values based on longitudinal (driving direction) and lateral directions as well as orientation localization errors. For $r@xm$, a prediction was considered correct if the directional error between the ground truth and prediction was within $x$ meters. For $r@x^\circ$, a prediction was considered correct if the orientation error was within $x^\circ$.
The evaluation metrics include the mean and median errors of location $(x, y)$ and orientation $(\theta)$ between the ground truth and prediction, where lower values indicate better performance. We also adopt the metric from SliceMatch~\cite{lentsch2023slicematch}, which reports recall based on longitudinal, lateral and orientation accuracy. For $r@xm$, a prediction is correct if the positional error is within $x$ meters, and for $r@x^\circ$, if the orientation error is within $x^\circ$.

%

% The quantitative results indicate that our method consistently achieves the best performance (highlighted in red) or the second-best performance (highlighted in blue) in most cases. This suggests that our method is robust in estimating poses across various sensor configurations, offsets, rotations, and environments. The effectiveness of our method is highlighted in Tab.~\ref{tab:exp_ford}, the experiment with the smaller dataset. Ours showed the best performance, with a large margin compared to the second-best. Other methods failed to estimate orientation, but our method achieves stable pose estimation with low orientation errors under limited ground image fields of view and arbitrary rotation, with errors less than $10^\circ$.

% The quantitative results indicate that our method consistently achieves the best performance (highlighted in red) or the second-best performance (highlighted in blue) in most cases. This suggests that our method is robust in estimating poses across various sensor configurations, offsets, rotations, and environments. Notably, when no prior for orientation is given ($\pm 180^\circ$ prior), our method achieves significantly higher accuracy compared to others. This is highlighted in Tab.~\ref{tab:exp_ford}, the experiment with the smaller dataset. Ours showed the best performance, with a large margin compared to the second-best. While other methods failed to estimate orientation, ours achieved stable pose estimation  with errors less than $10^\circ$ even under limited ground image fields of view.
The quantitative results show that the proposed method consistently achieves the best (red) or second-best (blue) performance across most settings, demonstrating strong robustness under various sensor configurations, offsets, rotations, and environmental conditions. When a small orientation prior ($\pm10^\circ$) is available, our method maintains competitive accuracy with only a minor gap from the best-performing cases. In contrast, when the orientation prior is removed ($\pm180^\circ$), the proposed method significantly outperforms all baselines, indicating strong robustness to large orientation uncertainty. This trend is particularly evident in Tab.~\ref{tab:exp_ford}, where our approach achieves the best performance by a large margin despite the relatively smaller dataset. Even in scenarios where other methods struggle to estimate orientation, our model maintains stable pose predictions with errors below $10^\circ$, despite the limited field of view of the ground images. Since orientation priors are often unavailable in real-world cross-view geo-localization scenarios, these results suggest that the proposed method has strong practical applicability.

The visualization of cross-view pose estimation results is shown in Fig.~\ref{fig:exp_pe_kitti}. Our method estimated poses closest to the ground truth in both location and orientation. Particularly, in the results of Fig.~\ref{fig:exp_pe_kitti}a, our method maintained accurate pose estimation even under a bridge where direct visibility from the aerial view is occluded. This qualitative result highlights the potential of the proposed method for robust cross-view geo-localization in challenging or GPS-denied environments.

\subsection{Ablation Study}

To validate the effectiveness of the proposed architecture, we conduct ablation studies on two key components: (1) the token configuration in the transformer encoder and (2) the bidirectional structure of the pose decoder. For the VIGOR dataset, only the NewYork and SanFrancisco subsets are used to conduct both same-area and cross-area evaluations.

First, we analyze whether the global retrieval and spatial localization representations are effectively disentangled in the dual-token encoder in Tab.~\ref{tab:exp_ablation_vigor}. We swap the tokens during inference to examine their roles. Since both tokens share the same ViT backbone, they exchange information through self-attention and thus retain a rich shared representation. Nevertheless, a small but consistent performance gap appears after swapping, indicating task specialization: retrieval relies on translation-invariant semantics, while pose estimation requires translation-variant spatial structure. This suggests that separating the roles across two tokens alleviates the multi-task bottleneck of a single representation.

Second, we evaluate the effect of the proposed two-way pose decoder by comparing it with a one-way cross-attention variant (Tab.~\ref{tab:exp_ablation_kitti_decoder}). Under a strong orientation prior ($10^\circ$), where cross-view matching is less ambiguous, the one-way variant occasionally performs comparably. However, when the prior is relaxed to $180^\circ$, the one-way design suffers from significant performance degradation. In contrast, the proposed two-way decoder maintains stable performance and consistently outperforms the baseline. This robustness is observed in both same-area and cross-area evaluations, demonstrating the advantage of bidirectional refinement under large viewpoint discrepancies and severe domain gaps.

\begin{table*}[t!]
\caption{\textbf{Token disentanglement analysis on cross-view geo-localization using VIGOR dataset}. $\text{T}_{cls}$ is the class token and $\text{T}_{pos}$ is the pose token.}
\centering
\vspace*{-0.35cm}
\begin{adjustbox}{width=0.6\linewidth}
\begin{tabular}{|c|c|cccc|cc|}
\hline 
\multirow{2}{*}{Area} & \multirow{2}{*}{Variant} & \multicolumn{4}{c|}{$(\uparrow)$ Image retrieval $(\%)$} & \multicolumn{2}{c|}{$(\downarrow)$ Pose estimation $(m)$}                                \\ \cline{3-8} 
                       &  & $R@1$         & $R@5$        & $R@10$        & $R@1\%$        & Mean & \multicolumn{1}{c|}{Median}  \\ \hline \hline
\multirow{2}{*}{same}
    & $\text{T}_{cls}$ & \textbf{41.39} & \textbf{68.93} & \textbf{77.52} & \textbf{98.5} & 5.24 & 4.12 \\ 
    & $\text{T}_{pos}$ & 40.88 & 68.25 & 76.96 & 98.44 & \textbf{5.17} & \textbf{4.07} \\ \hline

\multirow{2}{*}{cross}
    & $\text{T}_{cls}$ & \textbf{11.8} & \textbf{27.14} & \textbf{35.93} & \textbf{79.81} & 6.24 & 4.91 \\ 
    & $\text{T}_{pos}$ & 11.71 & 27.02 & 35.97 & 80.13 & \textbf{6.07} & \textbf{4.75} \\ \hline

\end{tabular}
\end{adjustbox}

\label{tab:exp_ablation_vigor}
\end{table*}
\begin{table*}[t!]
\caption{\textbf{Ablation study on cross-view pose estimation decoder architecture using KITTI dataset}}
\vspace*{-0.35cm}
\centering
\begin{adjustbox}{width=1\linewidth}
\begin{tabular}{|c|c|c|cc|cc|cc|cc|cc|}
\hline
\multirow{2}{*}{Area} & \multirow{2}{*}{Prior} & \multirow{2}{*}{Variant} & \multicolumn{2}{c|}{$(\downarrow)$ Location $(m)$} & \multicolumn{2}{c|}{$(\uparrow)$ Latitude $(\%)$} & \multicolumn{2}{c|}{$(\uparrow)$ Longitude $(\%)$} & \multicolumn{2}{c|}{$(\downarrow)$ Orientation $(^\circ)$} & \multicolumn{2}{c|}{$(\uparrow)$ Orientation $(\%)$} \\ \cline{4-13} 
 &  &  & Mean & Median & $R@1m$ & $R@5m$ & $R@1m$ & $R@5m$ & Mean & Median & $R@1^\circ$ & $R@5^\circ$ \\ \hline \hline

\multirow{2}{*}{same} & \multirow{2}{*}{$\pm 10^\circ$} 
 & One-way & 5.09 & 3.29 & 76.89 & 99.15 & 18.87 & 65.04 & 1.01 & 0.55 & 71.8 & 97.03 \\
 & & \textbf{Two-way (Ours)} & \textbf{2.02} & \textbf{1.38} & \textbf{88.74} &  \textbf{99.81} & \textbf{42.35} &  \textbf{94.04} & \textbf{0.69} & \textbf{0.44} & \textbf{80.28} & \textbf{99.55} \\ \hline

\multirow{2}{*}{same} & \multirow{2}{*}{$\pm 180^\circ$} 
 & One-way & 10.07 & 9.25 & 61.97 & 85.34 & 52.9 & 68.83 & 6.72 & 2.5 & 22.9 & 69.47 \\
 & & \textbf{Two-way (Ours)} & \textbf{8.26} & \textbf{6.81} &  \textbf{64.91} &  \textbf{89.85} &  \textbf{55.13} &  \textbf{74.48} &  \textbf{3.29} &  \textbf{1.31} &  \textbf{40.13} &  \textbf{90.19} \\ \hline
\hline
\multirow{2}{*}{cross} & \multirow{2}{*}{$\pm 10^\circ$} 
 & One-way & 10.14 & 7.48 & 38.35 & 88.27 & 9.33 & 40.33 & \textbf{1.43} & \textbf{0.9} & \textbf{53.83} & \textbf{95.97} \\
 & & \textbf{Two-way (Ours)} & \textbf{9.28} & \textbf{6.44} & \textbf{39.45} & \textbf{88.4} & \textbf{12.34} & \textbf{46.82} & 1.76 & 1.06 & 47.76 & 93.04 \\ \hline

\multirow{2}{*}{cross} & \multirow{2}{*}{$\pm 180^\circ$} 
 & One-way & 12.91 & 11.89 & 56.46 & 74.83 & 52.88 & 66.06 & 76.95 & 59.46 & 5.85 & 23.76 \\
 & & \textbf{Two-way (Ours)} & \textbf{12.54} &  \textbf{11.33} &  \textbf{56.93} &  \textbf{77.13} &  \textbf{53.47} &  \textbf{67.24} & \textbf{25.18} &  \textbf{4.48} & \textbf{15.45} &  \textbf{52.64} \\ \hline
\end{tabular}
\end{adjustbox}
\label{tab:exp_ablation_kitti_decoder}
\end{table*}

\section{Conclusion}
We proposed CIPER, a unified end-to-end transformer network for joint cross-view image retrieval and pose estimation, effectively integrating tasks previously handled by disjoint pipelines. By leveraging a dual-token shared encoder and a reciprocal cross-attention decoder, our approach mitigates the redundant feature extraction of cascaded methods and enables direct, stable 3-DoF localization. Extensive experiments demonstrated that our method achieves robust accuracy even under challenging conditions involving arbitrary orientations and limited fields of view, highlighting its strong potential for practical applications such as autonomous navigation in GPS-denied environments.

\bibliographystyle{splncs04}
\bibliography{egbib}

@inproceedings{toker2021coming,
  title={Coming down to earth: Satellite-to-street view synthesis for geo-localization},
  author={Toker, Aysim and Zhou, Qunjie and Maximov, Maxim and Leal-Taix{\'e}, Laura},
  booktitle={Proceedings of the IEEE/CVF Conference on Computer Vision and Pattern Recognition},
  pages={6488--6497},
  year={2021}
}

@inproceedings{hu2018cvm,
  title={Cvm-net: Cross-view matching network for image-based ground-to-aerial geo-localization},
  author={Hu, Sixing and Feng, Mengdan and Nguyen, Rang MH and Lee, Gim Hee},
  booktitle={Proceedings of the IEEE Conference on Computer Vision and Pattern Recognition},
  pages={7258--7267},
  year={2018}
}

@inproceedings{liu2019lending,
  title={Lending orientation to neural networks for cross-view geo-localization},
  author={Liu, Liu and Li, Hongdong},
  booktitle={Proceedings of the IEEE/CVF conference on computer vision and pattern recognition},
  pages={5624--5633},
  year={2019}
}

@article{shi2019spatial,
  title={Spatial-aware feature aggregation for image based cross-view geo-localization},
  author={Shi, Yujiao and Liu, Liu and Yu, Xin and Li, Hongdong},
  journal={Advances in Neural Information Processing Systems},
  volume={32},
  year={2019}
}

@inproceedings{shi2020optimal,
  title={Optimal feature transport for cross-view image geo-localization},
  author={Shi, Yujiao and Yu, Xin and Liu, Liu and Zhang, Tong and Li, Hongdong},
  booktitle={Proceedings of the AAAI Conference on Artificial Intelligence},
  volume={34},
  number={07},
  pages={11990--11997},
  year={2020}
}

@article{yang2021cross,
  title={Cross-view geo-localization with layer-to-layer transformer},
  author={Yang, Hongji and Lu, Xiufan and Zhu, Yingying},
  journal={Advances in Neural Information Processing Systems},
  volume={34},
  pages={29009--29020},
  year={2021}
}

@inproceedings{zhu2021vigor,
  title={Vigor: Cross-view image geo-localization beyond one-to-one retrieval},
  author={Zhu, Sijie and Yang, Taojiannan and Chen, Chen},
  booktitle={Proceedings of the IEEE/CVF Conference on Computer Vision and Pattern Recognition},
  pages={3640--3649},
  year={2021}
}

@inproceedings{regmi2019bridging,
  title={Bridging the domain gap for ground-to-aerial image matching},
  author={Regmi, Krishna and Shah, Mubarak},
  booktitle={Proceedings of the IEEE/CVF International Conference on Computer Vision},
  pages={470--479},
  year={2019}
}

@article{shi2022accurate,
  title={Accurate 3-DoF camera geo-localization via ground-to-satellite image matching},
  author={Shi, Yujiao and Yu, Xin and Liu, Liu and Campbell, Dylan and Koniusz, Piotr and Li, Hongdong},
  journal={IEEE Transactions on Pattern Analysis and Machine Intelligence},
  volume={45},
  number={3},
  pages={2682--2697},
  year={2022},
  publisher={IEEE}
}

@inproceedings{zhu2022transgeo,
  title={Transgeo: Transformer is all you need for cross-view image geo-localization},
  author={Zhu, Sijie and Shah, Mubarak and Chen, Chen},
  booktitle={Proceedings of the IEEE/CVF Conference on Computer Vision and Pattern Recognition},
  pages={1162--1171},
  year={2022}
}

@inproceedings{yan2022crossloc,
  title={Crossloc: Scalable aerial localization assisted by multimodal synthetic data},
  author={Yan, Qi and Zheng, Jianhao and Reding, Simon and Li, Shanci and Doytchinov, Iordan},
  booktitle={Proceedings of the IEEE/CVF Conference on Computer Vision and Pattern Recognition},
  pages={17358--17368},
  year={2022}
}

@article{kirillov2023segment,
  title={Segment anything},
  author={Kirillov, Alexander and Mintun, Eric and Ravi, Nikhila and Mao, Hanzi and Rolland, Chloe and Gustafson, Laura and Xiao, Tete and Whitehead, Spencer and Berg, Alexander C and Lo, Wan-Yen and others},
  journal={arXiv preprint arXiv:2304.02643},
  year={2023}
}

@inproceedings{shi2022beyond,
  title={Beyond cross-view image retrieval: Highly accurate vehicle localization using satellite image},
  author={Shi, Yujiao and Li, Hongdong},
  booktitle={Proceedings of the IEEE/CVF Conference on Computer Vision and Pattern Recognition},
  pages={17010--17020},
  year={2022}
}

@inproceedings{lentsch2023slicematch,
  title={SliceMatch: Geometry-guided Aggregation for Cross-View Pose Estimation},
  author={Lentsch, Ted and Xia, Zimin and Caesar, Holger and Kooij, Julian FP},
  booktitle={Proceedings of the IEEE/CVF Conference on Computer Vision and Pattern Recognition},
  pages={17225--17234},
  year={2023}
}

@article{geiger2013vision,
  title={Vision meets robotics: The kitti dataset},
  author={Geiger, Andreas and Lenz, Philip and Stiller, Christoph and Urtasun, Raquel},
  journal={The International Journal of Robotics Research},
  volume={32},
  number={11},
  pages={1231--1237},
  year={2013},
  publisher={Sage Publications Sage UK: London, England}
}

@article{agarwal2020ford,
  title={Ford multi-AV seasonal dataset},
  author={Agarwal, Siddharth and Vora, Ankit and Pandey, Gaurav and Williams, Wayne and Kourous, Helen and McBride, James},
  journal={The International Journal of Robotics Research},
  volume={39},
  number={12},
  pages={1367--1376},
  year={2020},
  publisher={SAGE Publications Sage UK: London, England}
}

@inproceedings{shi2023boosting,
  title={Boosting 3-DoF Ground-to-Satellite Camera Localization Accuracy via Geometry-Guided Cross-View Transformer},
  author={Shi, Yujiao and Wu, Fei and Perincherry, Akhil and Vora, Ankit and Li, Hongdong},
  booktitle={Proceedings of the IEEE/CVF International Conference on Computer Vision},
  pages={21516--21526},
  year={2023}
}

@inproceedings{xia2022visual,
  title={Visual cross-view metric localization with dense uncertainty estimates},
  author={Xia, Zimin and Booij, Olaf and Manfredi, Marco and Kooij, Julian FP},
  booktitle={European Conference on Computer Vision},
  pages={90--106},
  year={2022},
  organization={Springer}
}

@article{dosovitskiy2020image,
  title={An image is worth 16x16 words: Transformers for image recognition at scale},
  author={Dosovitskiy, Alexey and Beyer, Lucas and Kolesnikov, Alexander and Weissenborn, Dirk and Zhai, Xiaohua and Unterthiner, Thomas and Dehghani, Mostafa and Minderer, Matthias and Heigold, Georg and Gelly, Sylvain and others},
  journal={arXiv preprint arXiv:2010.11929},
  year={2020}
}

@inproceedings{carion2020end,
  title={End-to-end object detection with transformers},
  author={Carion, Nicolas and Massa, Francisco and Synnaeve, Gabriel and Usunier, Nicolas and Kirillov, Alexander and Zagoruyko, Sergey},
  booktitle={European conference on computer vision},
  pages={213--229},
  year={2020},
  organization={Springer}
}

@inproceedings{koch2015siamese,
  title={Siamese neural networks for one-shot image recognition},
  author={Koch, Gregory and Zemel, Richard and Salakhutdinov, Ruslan and others},
  booktitle={ICML deep learning workshop},
  volume={2},
  number={1},
  pages={1--30},
  year={2015},
  organization={Lille}
}

@inproceedings{arandjelovic2016netvlad,
  title={NetVLAD: CNN architecture for weakly supervised place recognition},
  author={Arandjelovic, Relja and Gronat, Petr and Torii, Akihiko and Pajdla, Tomas and Sivic, Josef},
  booktitle={Proceedings of the IEEE conference on computer vision and pattern recognition},
  pages={5297--5307},
  year={2016}
}

@inproceedings{wang2023pureACL,
  title={View Consistent Purification for Accurate Cross-View Localization},
  author={Wang, Shan and Zhang, Yanhao and Perincherry, Akhil and Vora, Ankit and Li, Hongdong},
  booktitle={Proceedings of the IEEE/CVF International Conference on Computer Vision},
  pages={8197--8206},
  year={2023}
  }
\end{document}